\documentclass[letterpaper, 10 pt, conference]{ieeeconf}  

\usepackage{times}
\usepackage{epsfig}
\usepackage{graphicx}
\usepackage{amsmath}
\usepackage{amssymb}
\usepackage{graphics} 
\usepackage{subfigure}
\usepackage{bm}
\usepackage{helvet}  
\usepackage{courier} 
\usepackage{url}     
\usepackage{algorithm}
\usepackage{algorithmic}
\usepackage{diagbox}
\usepackage{multicol,multirow}
\usepackage{subfigure}
\usepackage{cite}
\usepackage{listings}
\usepackage{xcolor}
\usepackage[american]{babel}
\usepackage{multirow}
\usepackage{soul}

\newcommand{\todo}[1]{{\textcolor{black}{#1}}}

\IEEEoverridecommandlockouts                              

\title{\LARGE \bf
Pose Refinement Graph Convolutional Network for Skeleton-based Action Recognition
}

\author{Shijie Li$^*$, Jinhui Yi$^*$, Yazan Abu Farha and Juergen Gall
\thanks{*These authors contributed equally to this work.}
}

\begin{document}

\maketitle
\thispagestyle{empty}
\pagestyle{empty}

\begin{abstract}
With the advances in capturing 2D or 3D skeleton data, skeleton-based action recognition has received an increasing interest over the last years. As skeleton data is commonly represented by graphs, graph convolutional networks have been proposed for this task. While current graph convolutional networks accurately recognize actions, they are too expensive for robotics applications where limited computational resources are available.
In this paper, we therefore propose a highly efficient graph convolutional network that addresses the limitations of previous works. This is achieved by a parallel structure that gradually fuses motion and spatial information and by reducing the temporal resolution as early as possible. Furthermore, we explicitly address the issue that human poses can contain errors. To this end, the network first refines the poses before they are further processed to recognize the action. We therefore call the network Pose Refinement Graph Convolutional Network. Compared to other  graph convolutional networks, our network requires 86\%-93\% less parameters and reduces the floating point operations by 89\%-96\% while achieving a comparable accuracy. It therefore provides a much better trade-off between accuracy, memory footprint and processing time, which makes it suitable for robotics applications.

\end{abstract}

\begin{figure*}[t]
    \centering
    \includegraphics[width=\linewidth]{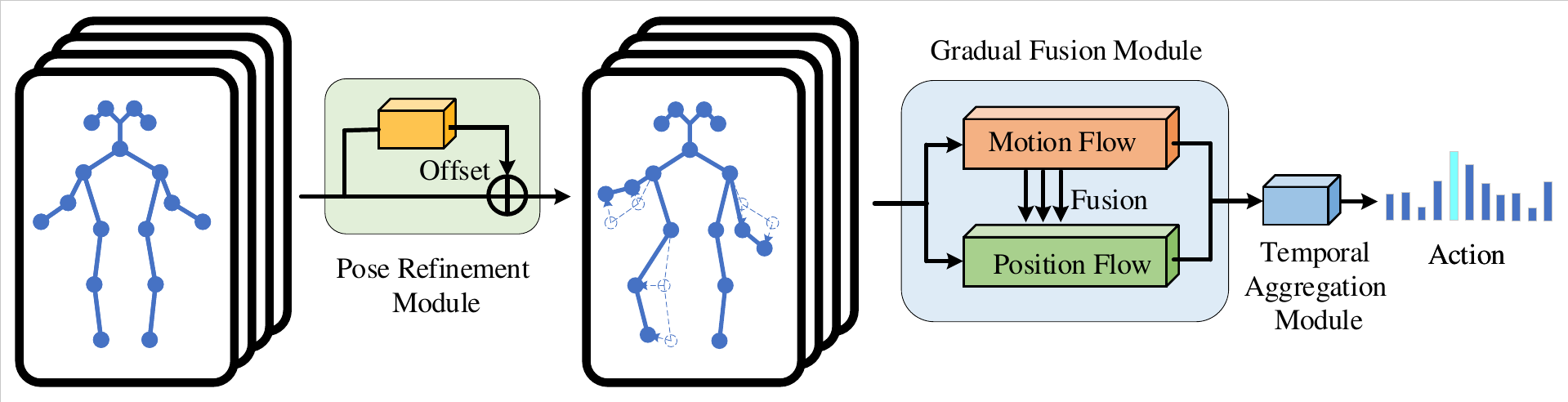}
    \caption{\todo{Pose Refinement Graph Convolutional Network.
    The input skeleton sequences are first passed through a pose refinement module to reduce the impact of errors in the skeleton data. 
    Then the refined skeleton sequences are fed into the gradual fusion module consisting of a motion-flow-branch and a position-flow-branch for fusing position and motion information.
    The position flow aggregates spatial information of skeleton joints at each time step whereas the motion flow captures the long-range temporal dependencies.
    Finally, the temporal aggregation module aggregates the information over time and predicts the action class probabilities.}}
    \label{fig:arch}
    \vspace{-5mm}
\end{figure*}

\section{INTRODUCTION}
Action recognition has received an increasing interest in recent years due to its importance for a broad range of applications such as video surveillance, gesture recognition and human-robot interaction. Although deep learning models are very popular for recognizing activities in videos~\cite{simonyan2014two, carreira2017quo,li2020ms}, these approaches are computationally expensive and cannot be used on mobile systems with limited computational resources. To alleviate this problem, skeleton data can be used for action recognition~\cite{vemulapalli2014human, fernando2015modeling}. 
In contrast to video-based approaches, skeleton-based action recognition models require much less computational resources and several approaches based on convolutional neural networks and recurrent neural networks have been proposed~\cite{ke2017new, zhang2017view, cao2018skeleton}. 

Despite the success of the previous approaches, their performance was limited as they did not consider the intrinsic differences between video data and skeleton data. While video frames have a grid structure where standard 2D or 3D convolutional neural networks can be applied, skeleton data is commonly represented by graphs. The spatial-temporal graph convolutional network (ST-GCN) \cite{yan2018spatial} therefore models skeleton sequences as spatial-temporal graphs and uses graph convolutions. Inspired by this work, further improvements have been proposed to increase the accuracy~\cite{li2019actional,shi2019two}. While the variants of graph convolutional networks achieve very good results in terms of accuracy, the improvements come at a high increase of memory consumption and computational cost. Indeed, \cite{shi2019two} has more than twice the number of parameters than ST-GCN and \todo{increases the number of floating point operations (FLOP) for inference} by a factor of around 2.5. This prevents the application of these networks within the robotics domain.

In this work, we therefore propose a graph convolutional network that achieves a higher action recognition accuracy compared to ST-GCN, but that is much more efficient. Instead of increasing the size of the model, we present a new architecture that requires 7 times less parameters than ST-GCN and reduces the \todo{FLOP} by a factor of 9. This is achieved by treating the temporal and spatial relations in a different way. To this end, we combine temporal and graph convolutions. While the graph convolutions focus on the spatial relations, the temporal convolutions aggregate the temporal information. Whereas in a sequential model the temporal and graph convolutions follow each other, we propose to separate them in a parallel structure and gradually fuse them as part of the gradual fusion module (GFM) as illustrated in Fig.~\ref{fig:arch}. In contrast to a spatial-temporal graph, the temporal dimension gets early reduced within the network and not only at the end, which results in a very efficient and compact network.

As a second contribution, we address the problem that the estimated 2D or 3D human poses can be inaccurate. Although many approaches for human pose estimation from RGB or depth sensors exist and some of them can be deployed on robotics platforms, pose estimation errors occur due to the limited view or occlusions. Previous works on skeleton-based action recognition did not address this issue explicitly, assuming that the networks deal with pose estimation errors implicitly. In this work, we propose to add a module that refines the human poses by taking the spatial and temporal information into account. 
The so-called pose refinement module (PRM) estimates offsets for each joint and refines the poses by the offsets. As shown in Fig.~\ref{fig:arch}, the pose refinement module first refines the poses and the gradual fusion module with an additional temporal aggregation estimates then the action class probabilities. The entire network is trained only with an action classification loss such that no additional supervision is required. We therefore call the approach pose refinement graph convolutional network (PR-GCN).        




Our contribution is thus two folded:
\begin{itemize}
    \item We propose the pose refinement graph convolutional network (PR-GCN), which is a compact model for skeleton-based action recognition that gradually fuses position and motion information and provides a better trade-off between effectiveness and efficiency compared to the state-of-the-art.
    \item We introduce the pose refinement module that reduces the impact of pose estimation errors and further improves the accuracy at a very small increase of computational cost.
\end{itemize}
We evaluate the proposed approach\footnote{The source code is available at \url{https://github.com/sj-li/PR-GCN}.} on two very challenging action recognition datasets, namely Kinetics~\cite{kay2017kinetics} with estimated 2D human poses~\cite{yan2018spatial} and NTU RGB+D~\cite{shahroudy2016ntu} with estimated 3D human poses. Compared to other variants of graph convolutional networks, our method achieves a competitive accuracy but at a small fraction of the required computational resources. Compared to the state-of-the-art approach \cite{shi2019two}, the number of parameters are reduced by factor 14 and the computational cost by factor 22, which makes it suitable for robotics applications\cite{li2018direct,li2018structured,li2020multi}.

\section{Related Work}
\label{sec:rel}

\subsection{Skeleton-based action recognition}

\todo{Skeleton-based action recognition is an important research area that has received an increasing attention recently.
While earlier approaches use hand-crafted features for action recognition~\cite{YaoGG12,6751508,vemulapalli2014human,fernando2015modeling}, more recent works use data-driven methods based on convolutional neural networks (CNNs) or recurrent neural networks (RNNs).
As CNNs are applied on images with a regular grid structure, CNN-based methods represent the skeleton data as pseudo-images~\cite{ke2017new, kim2017interpretable,li2017skeleton}. 
On the contrary, RNN-based methods are usually proposed for sequential data. Hence, RNN-based methods model the skeleton data as a sequence of vectors, each of them representing the coordinates of the body joints~\cite{shahroudy2016ntu, liu2016spatio, zhang2017view, zheng2018skeleton, li2018independently, cao2018skeleton}. 
Despite the success of both RNNs and CNNs, their performance is limited as they do not explicitly capture the dependencies between the body joints. 
Some approaches \cite{yan2018spatial,li2019actional,shi2019two} therefore model the skeleton data as spatio-temporal graph structures and use graph convolutions instead of 2D convolutions. Graphs have also been used to model other relations for action recognition. For instance, spatio-temporal graphs \cite{liu2020learning} or scene graphs \cite{dreher2019learning} are used to model relations between objects. In \cite{weiexplainable}, the problem of recognizing activities that occur at the same time is addressed by learning spatio-temporal correlations of activities. There are also a few works that proposed lightweight networks for action recognition~\cite{liu2020gfnet,zhang2019eleatt,yang2019make}.
}


\subsection{Graph convolutional neural networks}

\todo{There are many works that focus on graph convolutional networks (GCNs). The graphs are usually constructed in the spatial or spectral domain \cite{shuman2013emerging, bruna2013spectral, henaff2015deep, niepert2016learning, atwood2016diffusion, kipf2017semi, duvenaud2015convolutional, defferrard2016convolutional}.
In the spatial domain \cite{duvenaud2015convolutional, niepert2016learning, hamilton2017inductive, monti2017geometric, kipf2018neural}, the convolution operation is directly applied on the graph vertices and their neighbors.
Whereas spectral methods perform the graph convolutions in the frequency domain with the help of the graph Fourier transform \cite{shuman2013emerging}. The spectral methods do not require any extra effort to extract locally connected regions from graphs at each convolutional step \cite{shuman2013emerging, defferrard2016convolutional, henaff2015deep, kipf2017semi}. Over the years, several improvements have been proposed. For instance, an attention mechanism has been integrated into graph convolutional networks in \cite{velivckovic2017graph}. In \cite{li2018adaptive}, the distance metrics are parameterized  so that the graph Laplacian itself becomes trainable. The work \cite{gao2018large} proposes sub-graph training in order to deal with very large graphs.}


\section{Pose Refinement Graph Convolutional Network}
\label{sec:pr-gcn}
The proposed pose refinement graph convolutional network (PR-GCN) is a very compact and efficient network for skeleton-based action recognition which makes it suitable for robotics applications\cite{liu2020refinedbox,li2020projected}. 
\todo{The architecture of the proposed network is illustrated in Fig.~\ref{fig:arch}. Pose estimation errors of the input sequence are first corrected by the pose refinement module. Examples of refined human poses are shown in Fig.~\ref{fig:vis_refine}. The refined skeleton sequence is then processed by a motion-flow-branch and a position-flow-branch that are gradually fused. Finally, the temporal aggregation module generates the final prediction. In the following, we describe the network in detail.}


\subsection{Graph Construction}



\begin{figure}[tbp]
    \centering
    \includegraphics[width=\linewidth]{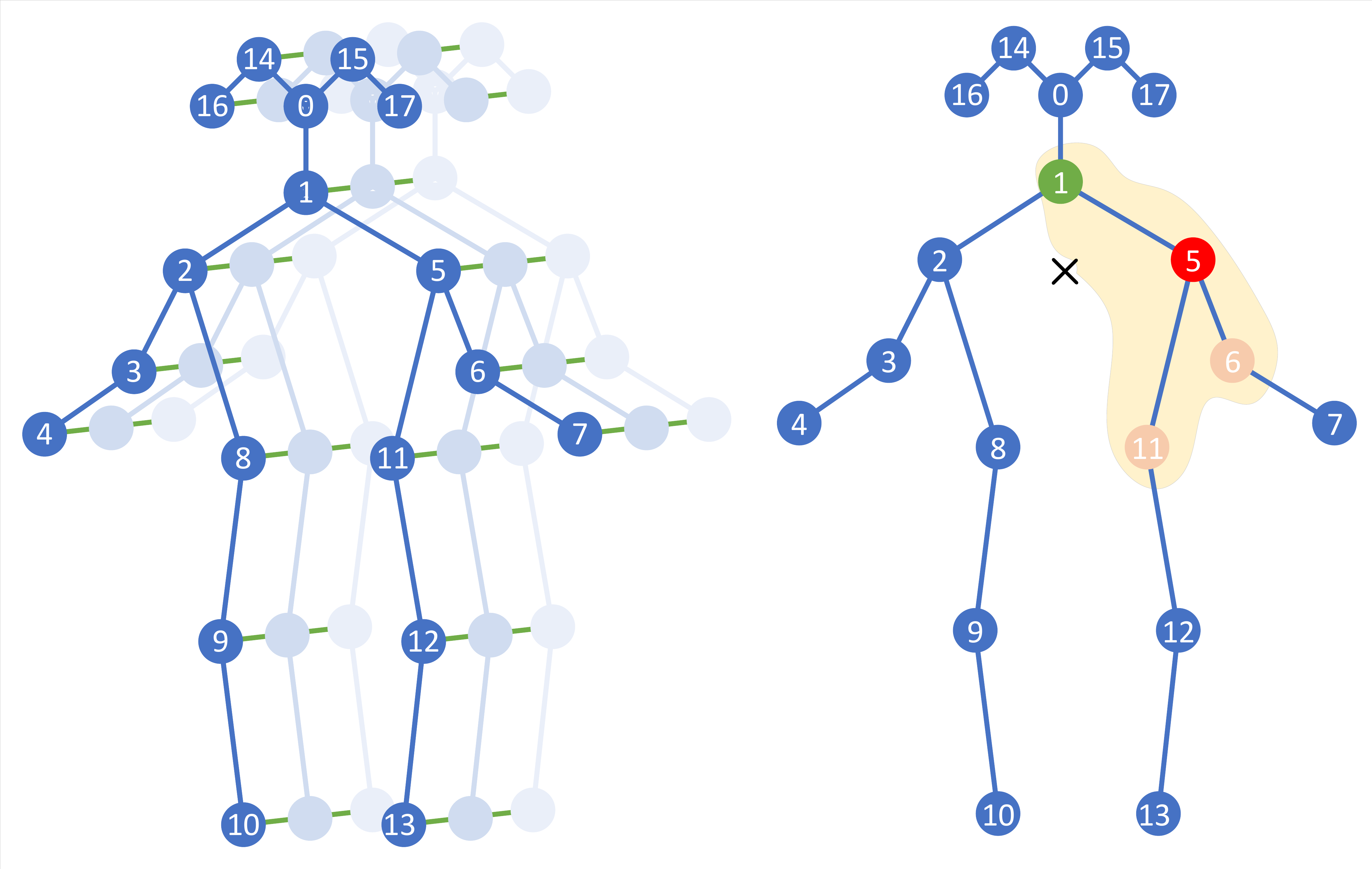}
    \caption{The left figure illustrates the spatio-temporal graph~\cite{yan2018spatial} for the 2D skeleton that is provided for Kinetics. The right figure shows the spatial neighbors of joint 5, which are grouped based on the vertex itself (red), the vertices that are closer to the center of gravity (green) and the vertices that are more far from the center of gravity (yellow). 
    }
    \label{fig:skeleton}
    \vspace{-2mm}
\end{figure}

Similar to ST-GCN~\cite{yan2018spatial}, each skeleton sequence is modeled as a spatial-temporal graph $\mathcal{G}(V, E)$ as shown in Fig.~\ref{fig:skeleton} (left).
The vertices $V$ denote body joints represented by their 2D or 3D joint coordinates.
\todo{The edges $E$ comprise spatial and temporal edges.} Spatial edges $E_s$ represent connections between vertices at each frame whereas temporal edges $E_t$ connect the same body joint between two adjacent frames. 

\subsection{Basic Operations}\label{sec:basic}
\todo{Since PR-GCN includes graph convolution and temporal convolution layers, we briefly describe them first.}

\subsubsection{Graph Convolution Layer}


The graph convolutions~\cite{yan2018spatial} operate on a pre-defined graph structure.
In the spatial dimension, the graph convolution for each vertex $v_i$ is formulated as: 
\begin{equation}
    f_{out}(v_i)=\sum_{v_j \in \mathcal{B}_i} \frac{1}{Z_{ij}}f_{in}(v_j)\cdot w(l_i(v_j)),
    \label{eq:gcn_def}
\end{equation}
where $f(v)$ denotes the feature for vertex $v$ and $w$ is a weight function. $v_i$ is the target vertex and $\mathcal{B}_i$ is the set of neighbor vertices including $v_i$. In our implementation, $\mathcal{B}_i$ contains all 1-distance neighbors as it is illustrated in Fig.~\ref{fig:skeleton} (right). 
Since the number of vertices in $B_i$ varies based on the selection of $v_i$, \cite{yan2018spatial} introduced a mapping function $l_i$ that maps neighbor vertices into a set of predefined groups: the vertex itself ($\mathcal{G}_{i1}$), the vertices that are close to the center of gravity ($\mathcal{G}_{i2}$), and vertices that are far from the center of gravity ($\mathcal{G}_{i3}$) as shown in Fig.~\ref{fig:skeleton} (right). This means that the weighting function $w$ generates three different weights, one for each group.   
$Z_{ij}$ is the cardinality of $\mathcal{G}_{ik}$ that contains $v_j$. It is used to balance the contribution of each group.

\begin{figure}[t]
    \centering
    \includegraphics[width=0.47\linewidth]{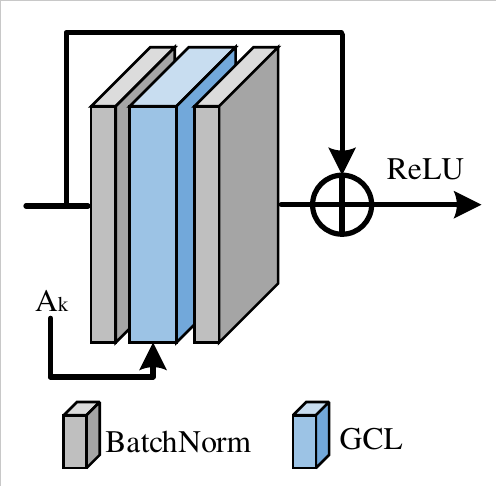}
    \includegraphics[width=0.51\linewidth]{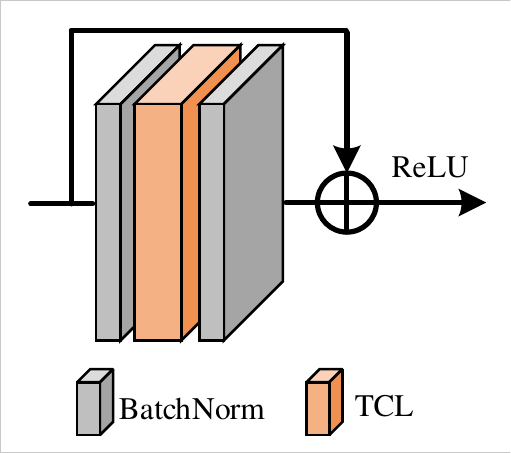} 
    \caption{Graph convolution layer (left) and temporal convolution layer (right).}
    \label{fig:TCM}
    \vspace{-2mm}
\end{figure}

The input skeleton sequence is organized as a tensor of size $(C, T, N)$, where $C$ is the number of channels, $T$ is the sequence length and $N$ denotes the number of vertices.
Hence, \eqref{eq:gcn_def} is transformed into:
\begin{equation}
    f_{out} = \sum_{k=1}^{K_v}\mathbf{W}_k(\mathbf{f}_{in}\mathbf{A}_k)\odot \mathbf{M}_k,
\end{equation}
\todo{where $K_v=3$ is the kernel size of the spatial dimension. The matrix $\mathbf{A}_k$ is defined by $\mathbf{A}_k = \mathbf{\Lambda}_k^{-\frac{1}{2}}\mathbf{\mathbf{\bar{A}}_k}\mathbf{\Lambda}_k^{-\frac{1}{2}}$ where $\mathbf{\bar{A}}_k$ indicates the connection of vertices and an element $\mathbf{\bar{A}}_{k}^{ij}$ is nonzero only if vertex $v_j \in B_i$.}
$\mathbf{\Lambda}_k^{ii} = \sum_{j}\mathbf{\bar{A}}_{k}^{ij} + \alpha$ is the normalized diagonal matrix and $\alpha$ is set as 0.001 to avoid division by zero.
$\mathbf{W}_k$ is a weight vector of a 1$\times$1 convolution operation and corresponds to $w$ in \eqref{eq:gcn_def}.
An attention map $\mathbf{M}_k$, which indicates the importance of each vertex, is applied on each vertex by an element-wise product $\odot$.
The graph convolution layer with additional batch normalization layers and a skip connection is shown in Fig.~\ref{fig:TCM} (left).  

\subsubsection{Temporal Convolution Layer}

For each vertex, there are only two connected vertices along the temporal dimension. The temporal convolution is a $K_t$ $\times$ 1 convolution where $K_t=3$ is the kernel size in the temporal dimension. We also use batch normalization and residual connections for the temporal convolution layers as shown in Fig.~\ref{fig:TCM} (right).



\begin{figure}[t]
    \centering
    \includegraphics[width=0.8\linewidth]{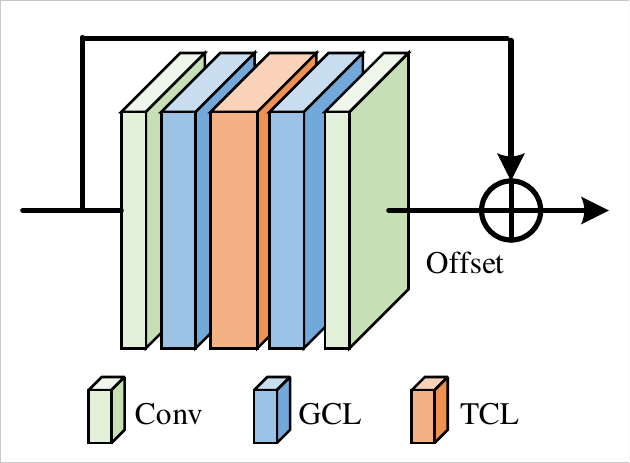}
    \caption{Pose refinement module.}
    \label{fig:pam}
\end{figure}

\subsection{Pose Refinement Module}
Due to partial visibility of the human body and inaccurate predictions from human pose estimation algorithms, the input skeleton data can contain errors, which influences the action recognition accuracy. To reduce this influence, we refine each joint position by the pose refinement module as shown in Fig.~\ref{fig:pam}. For pose refinement, we estimate offsets for each joint that are then added to the input poses.

In case of 3D skeleton data, we estimate the offset $(\Delta x, \Delta y, \Delta z)$ for each joint and add it to its 3D position  $(x, y, z)$. As shown in Fig.~\ref{fig:pam}, the offset is estimated by a combination of 1$\times$1 convolution layers, graph convolution layers and one temporal convolution layer. While the graph convolutional layers exploit the spatial relations of the joints to refine the poses, the temporal convolutional layer exploits temporal consistency. In case of 2D skeleton data, we use the $x$ and $y$ coordinates as well as the joint estimation confidence as input, which we obtain from the human pose estimation approach. The estimated 2D offset $(\Delta x, \Delta y)$ is then added to the 2D coordinates for each joint.


\begin{figure}[!htbp]
    \centering
    \includegraphics[width=\linewidth]{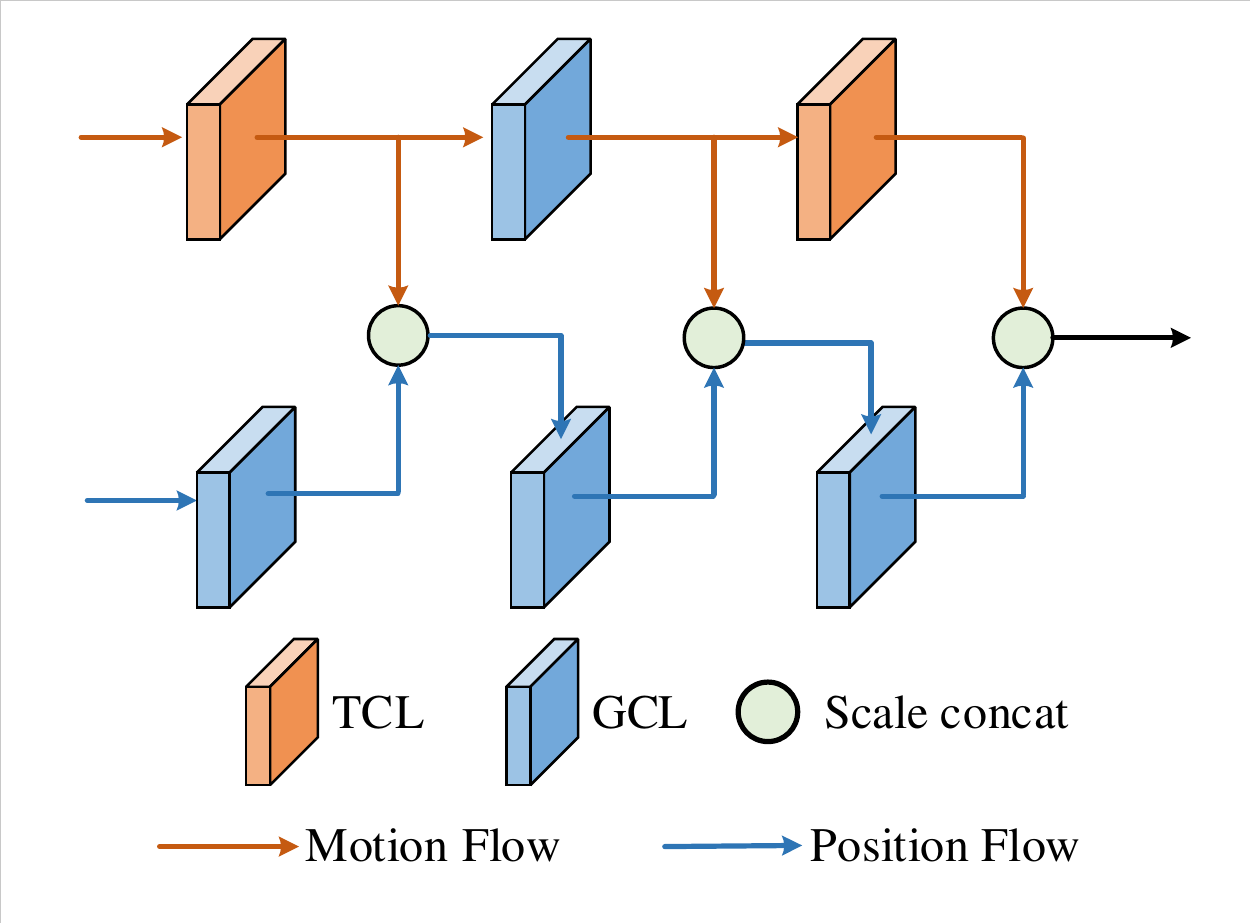}
    \caption{Gradual fusion module. Scale concat denotes the fusion operation shown in Fig.~\ref{fig:sc}.}
    \label{fig:gfm}
    \vspace{-2mm}
\end{figure}

\subsection{Gradual Fusion Module}
\begin{figure}[tb]
    \centering
    \includegraphics[width=0.8\linewidth]{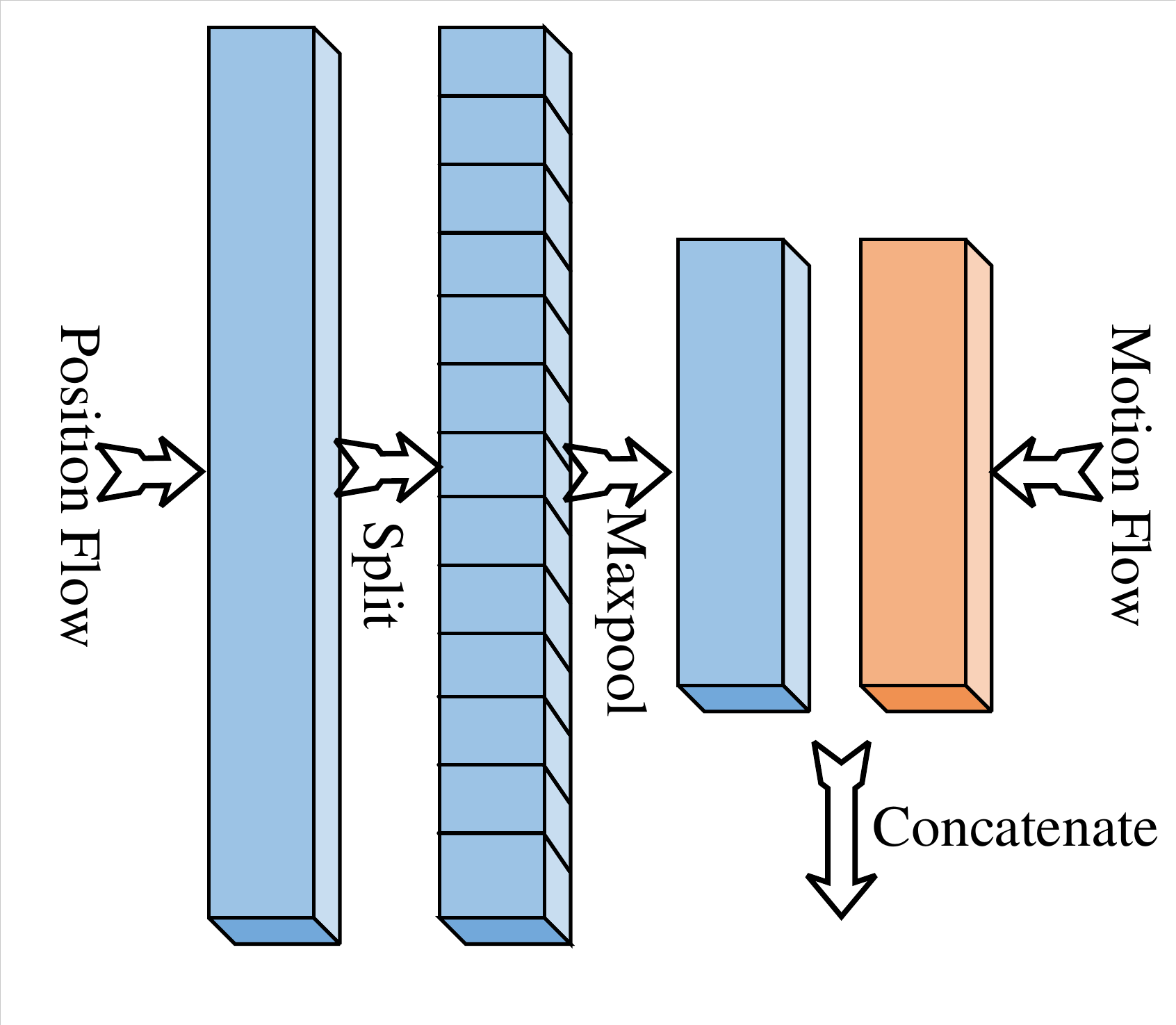}
    \caption{Fusion of features from the position and motion flow.}
    \label{fig:sc}
\end{figure}

Previous works like \cite{shi2019two} iterate between spatial and temporal graph convolutions. This is, however, not only very inefficient but we also show in Sec.~\ref{sec:exp} that this can even slightly decrease the accuracy when motion and spatial information are sequentially processed. Furthermore, the temporal edges do not model the full motion of the joints. Instead, we model the motion by temporal differences:
\begin{equation}
    M_{i, t} = P_{i, t} - P_{i, t-1}
\end{equation}
where $M_{i, t}$ is the motion of joint $i$ at time $t$, $P_{i, t}$ and $P_{i, t-1}$ are the positions of joint $i$ at time $t$ and $t-1$, respectively. As shown in Fig.~\ref{fig:gfm}, the motion flow takes $M$ as input and the position flow $P$. In order to include the spatial relations between the joints as additional information for the motion flow, we include one graph convolution layer in the motion flow. In order to capture long-term dependencies, we furthermore reduce the temporal resolution within the gradual fusion module by using stride 2 and 3 for the first and second temporal convolutional layer, respectively. The position flow is in parallel to the motion flow, but we gradually fuse the two flows as shown in Fig.~\ref{fig:gfm}. It is worth noting that the dimensions of the spatial features and the temporal features do not match due to the reduction of the temporal resolution in the motion flow. We therefore perform max-pooling over the spatial features along the temporal dimension such that the dimensions match at each fusion step. This is shown in Fig.~\ref{fig:sc}.

\begin{table*}[tb]
\centering
\resizebox{\linewidth}{!}
{%
\begin{tabular}{l|cc|cc|c|c|cc}
\hline
Methods & PRM & TAM  &   Top-1 (\%)   &    Top-5 (\%) & Params (M)  & \todo{GFLOP} & \todo{FPS (GPU)} & \todo{FPS (CPU)}\\
\hline
ST-GCN\cite{yan2018spatial} & & & 30.7 & 52.8 & 3.5 & 15.6 & 134.1 & 4.9 \\ 
AS-GCN\cite{li2019actional} & & & 34.8 & 56.5 & 5.0 & 27.0 & -&-\\
2s-AGCN\cite{shi2019two} & & & 36.1 & 58.7 & 7.1 & 38.5 & -&-\\
\hline
 \multirow{4}{*}{PR-GCN} & & & 29.3 & 51.8 & 0.3 & 1.3 & 570.1 & 20.3 \\
  & \checkmark & & 30.7 & 53.1 & 0.3 & 1.6 & 442.0 & 15.8 \\
  & & \checkmark & 33.2 & 55.5 & 0.5 &1.4 & 560.6 & 19.7\\ 
 & \checkmark & \checkmark & 33.6 & 56.1 & 0.5 & 1.7 & 433.9 & 15.7\\
\hline

\end{tabular}%
}
\caption{Ablation study on the impact of the pose refinement module (PRM) and the temporal aggregation module (TAM) on the Kinetics dataset. Compared to other graph convolutional networks, the proposed PR-GCN requires only a fraction \todo{of the number of parameters and giga floating point operations (GFLOP) for inference. The proposed approach also processes much more frames per second (FPS) on a GPU as well as on a CPU.}
}
\label{tab:ablation_module}
\vspace{-5mm}
\end{table*}



\begin{table}[tb]
\centering
\resizebox{0.8\linewidth}{!}
{%
\begin{tabular}{c|cccc}
\hline
Position & \checkmark & \checkmark & \checkmark & \checkmark \\
Motion   &  & \checkmark &  & \checkmark \\
\hline
Sequential & \checkmark & \checkmark &  &            \\
Parallel   &            &    & \checkmark & \checkmark \\
\hline
Top-1 (\%) & 32.9 & 32.2 &32.7 & 33.6\\
Top-5 (\%) & 55.1 & 55.0 & 55.2 & 56.1\\
\hline

\end{tabular}%
}
\caption{Ablation study for the gradual fusion module on the Kinetics dataset.}
\label{tab:ablation_gfm}
\end{table}



\begin{figure}[tb]
    \centering
    \includegraphics[width=\linewidth]{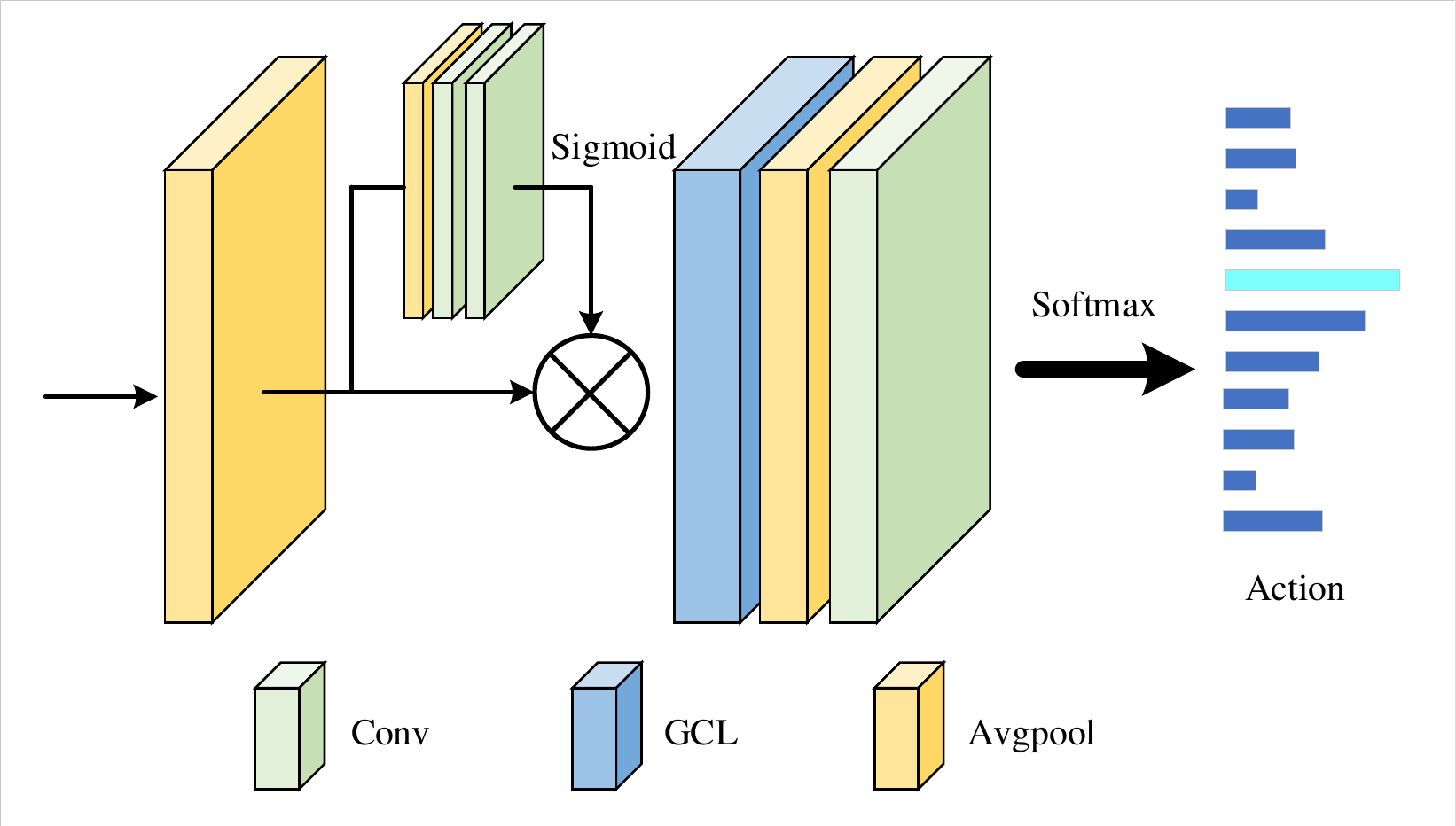}
    \caption{Temporal aggregation module.}
    \label{fig:tam}
    \vspace{-5mm}
\end{figure}

\subsection{Temporal Aggregation Module}
Finally, we aggregate the features to obtain the final class probabilities for each action class as shown in Fig.~\ref{fig:tam}. The fused features from the gradual fusion module are first fed into an average pooling layer across the temporal dimension. The features are then recalibrated according to the similarity among different channels. To this end, we scale each feature channel $f_{in}$ by a scalar value: 
\begin{equation*}
    f_{out} = f_{in}\cdot \sigma(\text{Conv}_o(\text{ReLU}(\text{Conv}_i(\text{Avg}(f_{in}))))).
\end{equation*}
While the average pooling layer (Avg) squeezes the input features $f_{in}$ into a vector, the $1\times1$ convolutions (Conv) learn the weight for each channel. The values are then mapped by the sigmoid function to values between zero and one. After the features are recalibrated, a graph convolution layer is applied to aggregate spatial information for the last time. Finally, the features are processed by an average pooling layer, a 
convolution layer, and a softmax function for predicting the probabilities of the action classes. 


\section{Experiments}
\label{sec:exp}

\subsection{Datasets \& Evaluation Metrics}
We evaluate our approach on two challenging large-scale human action datasets, namely Kinetics \cite{kay2017kinetics} and NTU RGB+D \cite{shahroudy2016ntu}.

\todo
{
\textbf{Kinetics} \cite{kay2017kinetics} contains video clips of 400 action classes retrieved from YouTube videos.
We use the human poses that are provided by \cite{yan2018spatial}\footnote{\url{https://github.com/kenziyuliu/st-gcn}}. The 2D human poses have been extracted using OpenPose~\cite{cao2017realtime}. To this end, the videos have been converted such that all videos have a resolution of 340$\times$256 pixels and a frame rate of 30 frames per second (FPS). The structure of the 2D skeleton is shown in Fig.~\ref{fig:skeleton}. The extracted skeleton data is split into a training set (240,000 clips) and a validation set (20,000 clips). As in \cite{yan2018spatial}, we report the top-1 and top-5 accuracy on the validation set.
}

\begin{figure*}[t]
\centering
\subfigure[Correcting pose errors]{
\begin{minipage}[t]{0.25\linewidth}
\centering
\includegraphics[trim={50 20 10 0}, width=\linewidth]{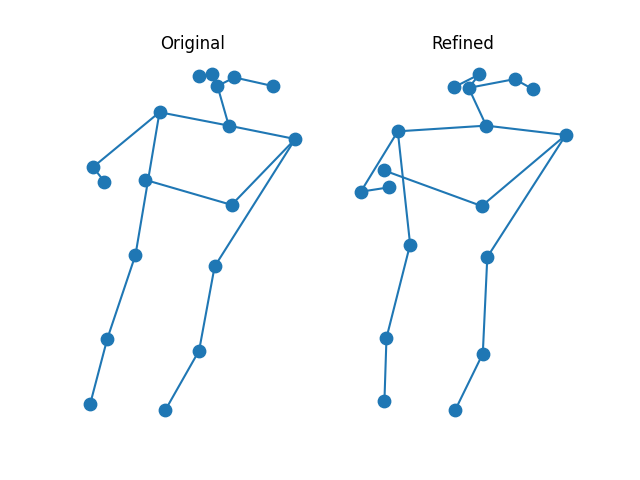}
\includegraphics[trim={50 20 10 0},width=\linewidth]{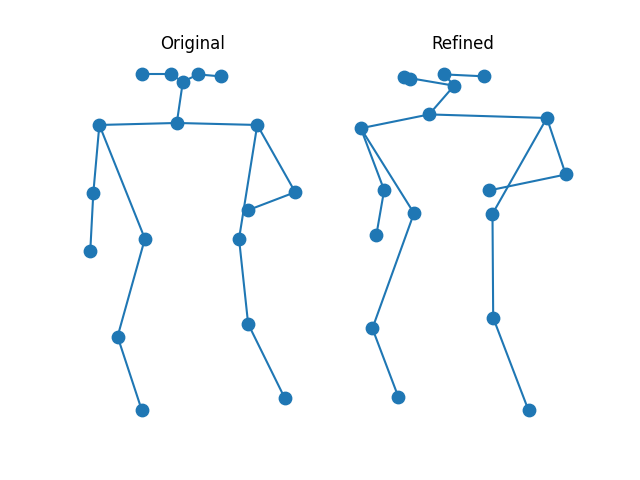}
\end{minipage}%
}%
\subfigure[Refining wrong poses]{
\begin{minipage}[t]{0.25\linewidth}
\centering
\includegraphics[trim={50 20 10 0},width=\linewidth]{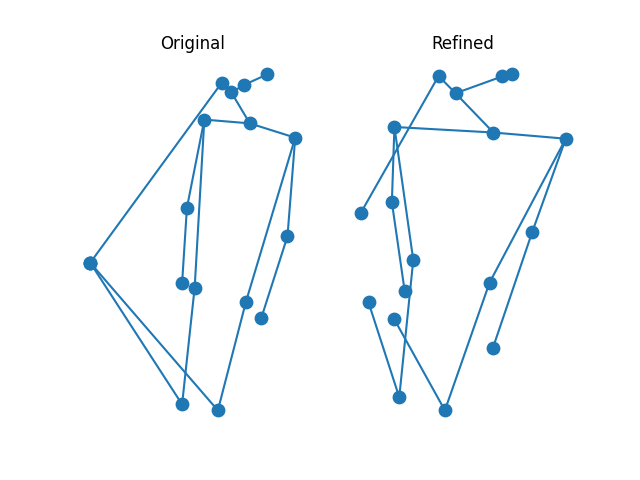}
\includegraphics[trim={50 20 10 0},width=\linewidth]{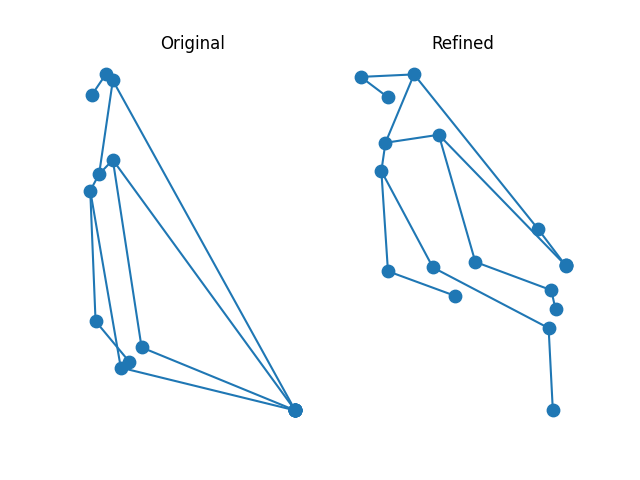}
\end{minipage}%
}%
\subfigure[Recovering missing joints]{
\begin{minipage}[t]{0.25\linewidth}
\centering
\includegraphics[trim={50 20 10 0},width=\linewidth]{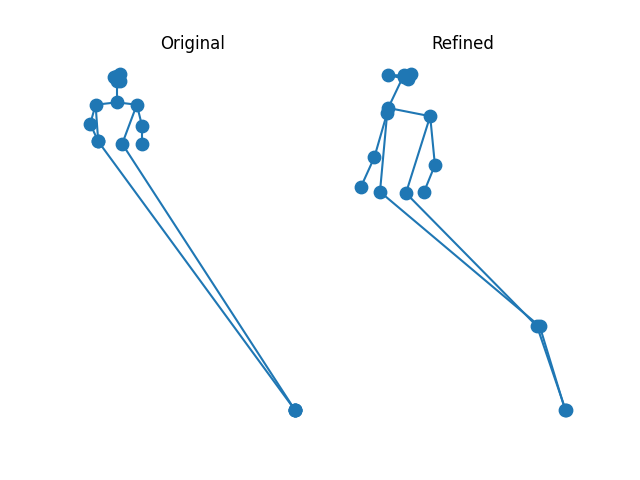}
\includegraphics[trim={50 20 10 0},width=\linewidth]{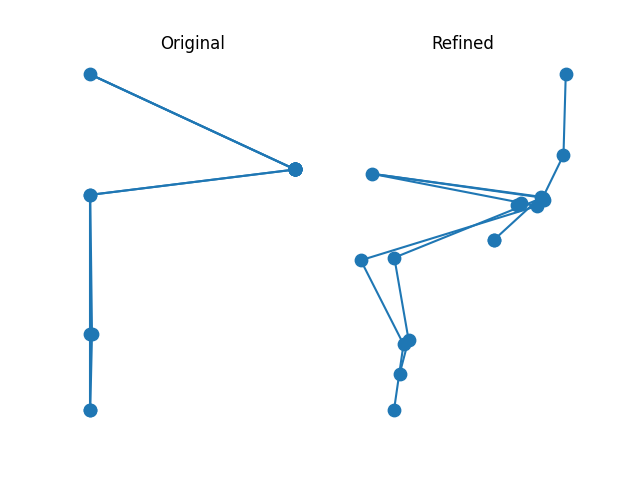}
\end{minipage}%
}%
\subfigure[Recovering missing poses]{
\begin{minipage}[t]{0.25\linewidth}
\centering
\includegraphics[trim={50 20 10 0},width=\linewidth]{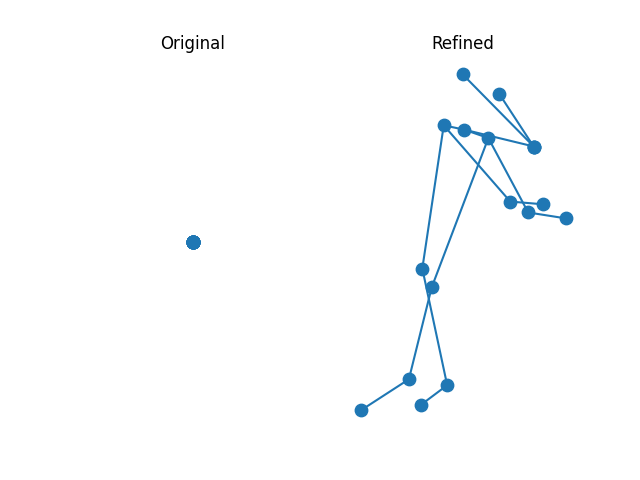}
\includegraphics[trim={50 20 10 0},width=\linewidth]{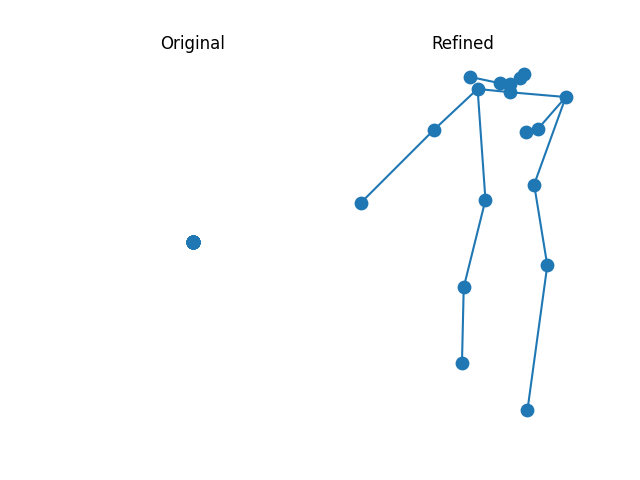}
\end{minipage}%
}%
\centering
\caption{\todo{Visualization of the pose refinement on the Kinetics dataset. Although we show only a single frame, the pose refinement takes the entire pose sequence into account. The examples show from left to right cases where (a) small pose errors are corrected, (b) wrong poses are refined, (c) missing joints are recovered, or (d) a missing pose is recovered.}}
\label{fig:vis_refine}
\end{figure*}

\textbf{NTU RGB+D} \cite{shahroudy2016ntu} contains 56,880 video clips with 60 actions that are performed by 40 persons, which are between 10 and 35 years old. The video clips have been recorded with multiple cameras and the 3D human poses are estimated by the Kinect v2 SDK. 
There are two evaluation protocols for this dataset \cite{shahroudy2016ntu}: 
\begin{itemize}
    \item Cross-subject (X-Sub): The videos are split into a training set (40,320 videos) and validation set (16,560) according to different actors.
    \item Cross-view (X-View): The videos are split according to different cameras. The training set contains 37,920 videos recorded by cameras 2 and 3, whereas the validation set includes 18,960 videos recorded by camera 1. 
\end{itemize}
We report top-1 accuracy for both protocols.
\todo{
All experiments are conducted using PyTorch on a machine equipped with an i7-5820K CPU (3.3 GHz), a GTX 1080TI GPU, and 16GB RAM.
}

\subsection{Implementation Details} 
Our model is trained with stochastic gradient descent with a learning rate of 0.01 and momentum of 0.9. 
The learning rate decays by 0.1 every 10 epochs apart from the first 10 epochs. 
The size of the input data of Kinetics is 300 frames. 
We apply data augmentation during training as \cite{yan2018spatial}, where 300 frames are randomly chosen from the input skeleton sequences and slightly disturbed with randomly chosen rotations and translations.

\subsection{Ablation Study}
We first examine the effectiveness of the pose refinement module (PRM) and the temporal aggregation module (TAM) in Tab.~\ref{tab:ablation_module}. If PRM is not included, the entire module is removed, i.e., we do not refine the poses for action recognition. If TAM is not included, we use only the last average pooling layer, 
convolution layer, and the softmax function of Fig.~\ref{fig:tam}. 

\todo
{
First, we evaluate our model without PRM and TAM. We can see from the table that our model achieves a slightly worse accuracy than ST-GCN~\cite{yan2018spatial}, but it has more than 11 times less parameters and it is more than 4 times faster on a GPU and on a CPU. Compared to ST-GCN, the number of giga floating point operations (GFLOP) are reduced by factor 12. 
}
One can also see that AS-GCN~\cite{li2019actional} and 2s-AGCN~\cite{shi2019two} increase the accuracy of ST-GCN by massively increasing the number of parameters and GFLOP.            

By adding the pose refinement module (PRM) and the temporal aggregation module (TAM) in Tab.~\ref{tab:ablation_module}, the accuracy increases while the number of parameters or runtime increases only slightly. By adding both PRM and TAM, we can see that both modules complement each other and achieve the best accuracy with an improvement of $4.3\%$ on the top-1 accuracy and top-5 accuracy. It is interesting to note that the pose refinement module only marginally increases the number of parameters but it slightly increases the runtime. Whereas the temporal aggregation module increases the parameters but only marginally the runtime. This is due to the fact that we have at the beginning of the network very few channels but the full temporal resolution and it is the other way around at the end of the network. Depending on the available memory or computational resources, one of the modules can be deactivated if necessary. While our approach outperforms ST-GCN~\cite{yan2018spatial} in terms of accuracy, memory footprint, and runtime, it achieves nearly the same accuracy as~\cite{li2019actional} but with only 10\% of the parameters and around 6\% of the GFLOP. Only the state-of-the-art method~\cite{shi2019two} achieves a higher accuracy, but at the cost of more than doubling the memory footprint and runtime of ST-GCN. \todo{Since GPUs are not always available for a robot system, we report the frames per seconds that are processed by ST-GCN~\cite{yan2018spatial} and our approach if they run on a GPU or CPU. In both cases, our approach is more than 3 times faster than~\cite{yan2018spatial}. It is also important to note that fast human pose estimation approaches like OpenPose process 15 or less frames per second on a robot platform, which is less than the 15.7 frames per second that are achieved by our approach on a CPU.}

\todo{In Fig.~\ref{fig:vis_refine}, we show a few examples that illustrate the effect of the pose refinement module. Besides of refining the estimated human poses, the module also recovers joints or poses that have not been detected either due to occlusion or due to a failure of the pose estimation approach.}

In a second ablation study, we explore the impact of the gradual fusion module. The results are shown in Tab.~\ref{tab:ablation_gfm}. To prove the effectiveness of our design, we tested two settings: a sequential architecture where we stack temporal convolutional layers and graph convolutional layers in a single flow and the parallel architecture shown in Fig.~\ref{fig:gfm}. For the sequential and parallel architecture, we also consider two cases. In case of using only position information, we use only the joint positions $P$ as input and in case of position and motion information, we use both $P$ and the motion of joints $M$. 

In the first two columns of Tab.~\ref{tab:ablation_gfm}, we evaluate the impact of using the additional motion information $M$ for a sequential architecture. In this case, we concatenate $P$ and $M$. The results show that the accuracy even slightly decreases if motion is added to a sequential architecture. When we compare the columns 1 and 3, we see that there is no substantial difference between a sequential and parallel architecture if only $P$ is used. In the parallel case, we use $P$ as input for both branches. Only if we use $P$ and $M$ for the parallel architecture as shown in Fig.~\ref{fig:gfm}, we observe an increase in accuracy as reported in the last column. This shows on the one hand that the additional motion information improves the accuracy and on the other hand that the proposed fusion module is essential for fusing the position and motion information.

\subsection{Comparison with State-of-the-Art}

We finally compare our method with current state-of-the-art methods for skeleton-based action recognition on the Kinetics and NTU RGB+D datasets. The results are shown in Tab.~\ref{tab:stoa_k} and Tab.~\ref{tab:stoa_n}.
The tables include methods with handcrafted features~\cite{fernando2015modeling}, RNN-based methods~\cite{shahroudy2016ntu}, CNN-based
methods~\cite{kim2017interpretable,yeh2019chirality} and GCN-based methods~\cite{yan2018spatial,li2019actional,shi2019two}. As shown in the tables, our method achieves an accuracy that is better or close to state-of-the-art methods. This shows that the proposed approach works very well for 2D human poses as well as 3D human poses. The approaches that achieve a higher accuracy use adaptive graph convolutions \cite{li2019actional} or adopt auxiliary strategies to further improve the accuracy like an ensemble of networks \cite{shi2019two}. This, however, makes these approaches demanding in terms of computational resources and difficult to deploy on robot platforms. As shown in Tab.~\ref{tab:ablation_module}, our network requires only 10\% of the parameters and around 6\% of the GFLOP compared to \cite{li2019actional} and only 7\% of the parameters and around 4\% of the GFLOP compared to \cite{shi2019two}.

\todo{
Furthermore, we also compare our method with other lightweight methods for action recognition on the NTU RGB-D dataset in Tab. \ref{tab:stoa_efficiency}. While GFNet \cite{liu2020gfnet} has less parameters and achieves a higher accuracy than ST-GCN \cite{yan2018spatial}, it requires more parameters than our approach and performs worse. Only EleAtt-GRU \cite{zhang2019eleatt} has less parameters than our approach, but its accuracy is even worse than ST-GCN.}
Our model achieves therefore a better trade-off between efficiency and accuracy compared to the state-of-the-art and is suitable for robotics applications with limited computational resources.

\begin{table}[tb]
\centering
\resizebox{0.85\linewidth}{!}
{%
\begin{tabular}{lccccc}
\hline
Methods     &    Top-1 (\%)   &    Top-5 (\%) \\
\hline
Feature Enc \cite{fernando2015modeling} &     14.9        &     25.8      \\
Deep LSTM \cite{shahroudy2016ntu}   &     16.4        &     35.3      \\
TCN \cite{kim2017interpretable}        &     20.3        &     40.0      \\
Conv \cite{yeh2019chirality} &     30.8        &     52.6      \\
Conv-Chiral \cite{yeh2019chirality} &     30.9        &     53.0      \\
ST-GCN \cite{yan2018spatial}     &     30.7        &     52.8      \\
AS-GCN\cite{li2019actional} & 34.8 & 56.5 \\
2s-AGCN\cite{shi2019two} & 36.1 & 58.7 \\
\hline
Ours        &     33.7   &     55.8      \\
\hline
\end{tabular}%
}
\caption{Comparison with state-of-the-art methods on the Kinetics dataset.}
\label{tab:stoa_k}
\vspace{-5mm}
\end{table}

\begin{table}[tb]
\centering
\resizebox{\linewidth}{!}
{%
\begin{tabular}{lccccc}
\hline
Methods     &  X-Sub (\%) & X-View (\%) \\
\hline
Lie Group\cite{vemulapalli2014human} & 50.1 & 82.8 \\
HBRNN\cite{du2015hierarchical} & 59.1 & 64.0 \\
Deep LSTM\cite{shahroudy2016ntu} & 60.7 & 67.3 \\
ST-LSTM\cite{liu2016spatio} & 69.2 & 77.7 \\
STA-LSTM\cite{song2017end} & 73.4 & 81.2 \\
VA-LSTM\cite{zhang2017view} & 79.2 & 87.7 \\
ARRN-LSTM\cite{zheng2018skeleton} & 81.8 & 89.6 \\
TCN\cite{kim2017interpretable} & 74.3 & 83.1 \\
Clips+CNN+MTLN\cite{ke2017new} & 79.6 & 84.8 \\
Synthesized CNN\cite{liu2017enhanced} & 80.0 & 87.2 \\
RGB+Skeleton\cite{fan2020context} & 84.2 & 89.3 \\
FO-GASTM\cite{li2019learning} & 82.8 & 90.1 \\
Bayesian GC-LSTM\cite{zhao2019bayesian} & 81.8 & 89.0 \\
GFNet \cite{liu2020gfnet} & 82.0 & 89.9 \\
EleAtt-GRU\cite{zhang2019eleatt} & 79.8 & 87.1 \\
ST-GCN\cite{yan2018spatial} & 81.5 & 88.3 \\
AS-GCN\cite{li2019actional} & 86.8 & 94.2 \\
2s-AGCN\cite{shi2019two} & 88.5 & 95.1\\
\hline
Ours        &     85.2   &     91.7     \\
\hline
\end{tabular}%
}
\caption{Comparison with state-of-the-art methods on the NTU RGB+D dataset.}
\label{tab:stoa_n}
\end{table}

\begin{table}[tb]
\centering
\resizebox{\linewidth}{!}
{%
\begin{tabular}{lcccc} 
\hline
Methods     &  X-Sub (\%) & X-View (\%) & Params (M) & GFLOP \\
\hline
ST-GCN \cite{yan2018spatial} & 81.5 & 88.3 & 3.5 & 15.6\\
GFNet \cite{liu2020gfnet} & 82.0 & 89.9 & 1.6 & 48.7 \\
EleAtt-GRU \cite{zhang2019eleatt} & 79.8 & 87.1 & 0.3 & -\\
\hline
Ours & 85.2 & 91.7 & 0.5 & 1.7 \\
\hline
\end{tabular}%
}
\caption{\todo{Comparison to other lightweight methods on the NTU RGB+D dataset.}}
\label{tab:stoa_efficiency}
\vspace{-1cm}
\end{table}

\section{CONCLUSIONS}
\label{sec:conclusion}

In this paper, we proposed a highly efficient model called Pose Refinement Graph Convolutional Network for 2D or 3D skeleton-based action recognition. It refines the human poses and gradually fuses motion and spatial information. Compared to previous graph convolutional networks, the proposed approach is very efficient in terms of memory footprint and runtime. It reduces the number of parameters by 86\%-93\% and the computational operations by 89\%-96\% while achieving a comparable accuracy. It therefore provides a much better trade-off between efficiency and accuracy and is thus suitable for robotics applications.




\section*{ACKNOWLEDGMENT}

The work has been financially supported by the Deutsche Forschungsgemeinschaft (DFG, German Research Foundation) under Germany's Excellence Strategy - EXC 2070 - 390732324, GA1927/4-2 (FOR 2535 Anticipating Human Behavior), and the ERC Starting Grant ARCA (677650).


\bibliographystyle{IEEEtran}
\bibliography{iros}

\begin{thebibliography}{10}
\providecommand{\url}[1]{#1}
\csname url@samestyle\endcsname
\providecommand{\newblock}{\relax}
\providecommand{\bibinfo}[2]{#2}
\providecommand{\BIBentrySTDinterwordspacing}{\spaceskip=0pt\relax}
\providecommand{\BIBentryALTinterwordstretchfactor}{4}
\providecommand{\BIBentryALTinterwordspacing}{\spaceskip=\fontdimen2\font plus
\BIBentryALTinterwordstretchfactor\fontdimen3\font minus
  \fontdimen4\font\relax}
\providecommand{\BIBforeignlanguage}[2]{{%
\expandafter\ifx\csname l@#1\endcsname\relax
\typeout{** WARNING: IEEEtran.bst: No hyphenation pattern has been}%
\typeout{** loaded for the language `#1'. Using the pattern for}%
\typeout{** the default language instead.}%
\else
\language=\csname l@#1\endcsname
\fi
#2}}
\providecommand{\BIBdecl}{\relax}
\BIBdecl

\bibitem{simonyan2014two}
K.~Simonyan and A.~Zisserman, ``Two-stream convolutional networks for action
  recognition in videos,'' in \emph{Advances in Neural Information Processing
  Systems}, 2014, pp. 568--576.

\bibitem{carreira2017quo}
J.~Carreira and A.~Zisserman, ``Quo vadis, action recognition? {A} new model
  and the kinetics dataset,'' in \emph{IEEE Conference on Computer Vision and
  Pattern Recognition}, 2017, pp. 4724--4733.

\bibitem{li2020ms}
S.-J. Li, Y.~AbuFarha, Y.~Liu, M.-M. Cheng, and J.~Gall, ``Ms-tcn++:
  Multi-stage temporal convolutional network for action segmentation,''
  \emph{IEEE Transactions on Pattern Analysis and Machine Intelligence}, 2020.

\bibitem{vemulapalli2014human}
R.~Vemulapalli, F.~Arrate, and R.~Chellappa, ``Human action recognition by
  representing 3d skeletons as points in a lie group,'' in \emph{IEEE
  Conference on Computer Vision and Pattern Recognition}, 2014, pp. 588--595.

\bibitem{fernando2015modeling}
B.~Fernando, E.~Gavves, J.~M. Oramas, A.~Ghodrati, and T.~Tuytelaars,
  ``Modeling video evolution for action recognition,'' in \emph{IEEE Conference
  on Computer Vision and Pattern Recognition}, 2015, pp. 5378--5387.

\bibitem{ke2017new}
Q.~Ke, M.~Bennamoun, S.~An, F.~Sohel, and F.~Boussaid, ``A new representation
  of skeleton sequences for 3d action recognition,'' in \emph{IEEE Conference
  on Computer Vision and Pattern Recognition}, 2017, pp. 3288--3297.

\bibitem{zhang2017view}
P.~Zhang, C.~Lan, J.~Xing, W.~Zeng, J.~Xue, and N.~Zheng, ``View adaptive
  recurrent neural networks for high performance human action recognition from
  skeleton data,'' in \emph{IEEE International Conference on Computer Vision},
  2017, pp. 2117--2126.

\bibitem{cao2018skeleton}
C.~Cao, C.~Lan, Y.~Zhang, W.~Zeng, H.~Lu, and Y.~Zhang, ``Skeleton-based action
  recognition with gated convolutional neural networks,'' \emph{IEEE
  Transactions on Circuits and Systems for Video Technology}, 2018.

\bibitem{yan2018spatial}
S.~Yan, Y.~Xiong, and D.~Lin, ``Spatial temporal graph convolutional networks
  for skeleton-based action recognition,'' in \emph{AAAI Conference on
  Artificial Intelligence}, 2018.

\bibitem{li2019actional}
M.~Li, S.~Chen, X.~Chen, Y.~Zhang, Y.~Wang, and Q.~Tian, ``Actional-structural
  graph convolutional networks for skeleton-based action recognition,'' in
  \emph{IEEE Conference on Computer Vision and Pattern Recognition}, 2019, pp.
  3595--3603.

\bibitem{shi2019two}
L.~Shi, Y.~Zhang, J.~Cheng, and H.~Lu, ``Two-stream adaptive graph
  convolutional networks for skeleton-based action recognition,'' in \emph{IEEE
  Conference on Computer Vision and Pattern Recognition}, 2019, pp.
  12\,026--12\,035.

\bibitem{kay2017kinetics}
W.~Kay, J.~Carreira, K.~Simonyan, B.~Zhang, C.~Hillier, S.~Vijayanarasimhan,
  F.~Viola, T.~Green, T.~Back, P.~Natsev \emph{et~al.}, ``The kinetics human
  action video dataset,'' \emph{arXiv preprint arXiv:1705.06950}, 2017.

\bibitem{shahroudy2016ntu}
A.~Shahroudy, J.~Liu, T.-T. Ng, and G.~Wang, ``{NTU RGB+D}: A large scale
  dataset for 3d human activity analysis,'' in \emph{IEEE Conference on
  Computer Vision and Pattern Recognition}, 2016, pp. 1010--1019.

\bibitem{li2018direct}
S.-J. Li, B.~Ren, Y.~Liu, M.-M. Cheng, D.~Frost, and V.~A. Prisacariu, ``Direct
  line guidance odometry,'' in \emph{2018 IEEE international conference on
  Robotics and automation (ICRA)}.\hskip 1em plus 0.5em minus 0.4em\relax IEEE,
  2018, pp. 1--7.

\bibitem{li2018structured}
S.-J. Li, M.-M. Cheng, Y.~Liu, S.-P. Lu, Y.~Wang, and V.~A. Prisacariu,
  ``Structured skip list: A compact data structure for 3d reconstruction,'' in
  \emph{2018 IEEE/RSJ International Conference on Intelligent Robots and
  Systems (IROS)}.\hskip 1em plus 0.5em minus 0.4em\relax IEEE, 2018, pp. 1--7.

\bibitem{li2020multi}
S.~Li, X.~Chen, Y.~Liu, D.~Dai, C.~Stachniss, and J.~Gall, ``Multi-scale
  interaction for real-time lidar data segmentation on an embedded platform,''
  \emph{arXiv preprint arXiv:2008.09162}, 2020.

\bibitem{YaoGG12}
A.~Yao, J.~Gall, and L.~Van~Gool, ``Coupled action recognition and pose
  estimation from multiple views,'' \emph{International Journal of Computer
  Vision}, vol. 100, no.~1, pp. 16--37, 2012.

\bibitem{6751508}
H.~{Jhuang}, J.~{Gall}, S.~{Zuffi}, C.~{Schmid}, and M.~J. {Black}, ``Towards
  understanding action recognition,'' in \emph{IEEE International Conference on
  Computer Vision}, 2013, pp. 3192--3199.

\bibitem{kim2017interpretable}
T.~S. Kim and A.~Reiter, ``Interpretable 3d human action analysis with temporal
  convolutional networks,'' in \emph{IEEE Conference on Computer Vision and
  Pattern Recognition Workshops}, 2017, pp. 1623--1631.

\bibitem{li2017skeleton}
C.~Li, Q.~Zhong, D.~Xie, and S.~Pu, ``Skeleton-based action recognition with
  convolutional neural networks,'' in \emph{IEEE International Conference on
  Multimedia \& Expo Workshops}, 2017, pp. 597--600.

\bibitem{liu2016spatio}
J.~Liu, A.~Shahroudy, D.~Xu, and G.~Wang, ``Spatio-temporal {LSTM} with trust
  gates for 3d human action recognition,'' in \emph{European Conference on
  Computer Vision}, 2016, pp. 816--833.

\bibitem{zheng2018skeleton}
W.~{Zheng}, L.~{Li}, Z.~{Zhang}, Y.~{Huang}, and L.~{Wang}, ``Relational
  network for skeleton-based action recognition,'' in \emph{IEEE International
  Conference on Multimedia and Expo}, 2019, pp. 826--831.

\bibitem{li2018independently}
S.~Li, W.~Li, C.~Cook, C.~Zhu, and Y.~Gao, ``Independently recurrent neural
  network ({IndRNN}): Building a longer and deeper {RNN},'' in \emph{IEEE
  Conference on Computer Vision and Pattern Recognition}, 2018, pp. 5457--5466.

\bibitem{liu2020learning}
Z.~Liu, Y.~Yao, Y.~Liu, Y.~Zhu, Z.~Tao, L.~Wang, and Y.~Feng, ``Learning
  dynamic spatio-temporal relations for human activity recognition,''
  \emph{IEEE Access}, vol.~8, pp. 130\,340--130\,352, 2020.

\bibitem{dreher2019learning}
C.~R.~G. {Dreher}, M.~{Wächter}, and T.~{Asfour}, ``Learning object-action
  relations from bimanual human demonstration using graph networks,''
  \emph{IEEE Robotics and Automation Letters}, vol.~5, no.~1, pp. 187--194,
  2020.

\bibitem{weiexplainable}
Y.~Wei, W.~Li, M.-C. Chang, H.~Jin, and S.~Lyu, ``Explainable and efficient
  sequential correlation network for 3d single person concurrent activity
  detection,'' in \emph{IEEE/RSJ International Conference on Intelligent Robots
  and Systems}, 2020.

\bibitem{liu2020gfnet}
H.~Liu, L.~Zhang, L.~Guan, and M.~Liu, ``{GFNet}: A lightweight group frame
  network for efficient human action recognition,'' in \emph{IEEE International
  Conference on Acoustics, Speech and Signal Processing}, 2020, pp. 2583--2587.

\bibitem{zhang2019eleatt}
P.~{Zhang}, J.~{Xue}, C.~{Lan}, W.~{Zeng}, Z.~{Gao}, and N.~{Zheng},
  ``{EleAtt-RNN}: Adding attentiveness to neurons in recurrent neural
  networks,'' \emph{IEEE Transactions on Image Processing}, vol.~29, pp.
  1061--1073, 2020.

\bibitem{yang2019make}
F.~Yang, Y.~Wu, S.~Sakti, and S.~Nakamura, ``Make skeleton-based action
  recognition model smaller, faster and better,'' in \emph{ACM Multimedia
  Asia}, 2019, pp. 1--6.

\bibitem{shuman2013emerging}
D.~I. Shuman, S.~K. Narang, P.~Frossard, A.~Ortega, and P.~Vandergheynst, ``The
  emerging field of signal processing on graphs: Extending high-dimensional
  data analysis to networks and other irregular domains,'' \emph{IEEE Signal
  Processing Magazine}, vol.~30, no.~3, pp. 83--98, 2013.

\bibitem{bruna2013spectral}
J.~Bruna, W.~Zaremba, A.~Szlam, and Y.~LeCun, ``Spectral networks and locally
  connected networks on graphs,'' in \emph{International Conference on Learning
  Representations}, 2014.

\bibitem{henaff2015deep}
M.~Henaff, J.~Bruna, and Y.~LeCun, ``Deep convolutional networks on
  graph-structured data,'' \emph{arXiv preprint arXiv:1506.05163}, 2015.

\bibitem{niepert2016learning}
M.~Niepert, M.~Ahmed, and K.~Kutzkov, ``Learning convolutional neural networks
  for graphs,'' in \emph{International Conference on Machine Learning}, 2016,
  pp. 2014--2023.

\bibitem{atwood2016diffusion}
J.~Atwood and D.~Towsley, ``Diffusion-convolutional neural networks,'' in
  \emph{Advances in Neural Information Processing Systems}, 2016, pp.
  1993--2001.

\bibitem{kipf2017semi}
T.~N. Kipf and M.~Welling, ``Semi-supervised classification with graph
  convolutional networks,'' in \emph{International Conference on Learning
  Representations}, 2017.

\bibitem{duvenaud2015convolutional}
D.~K. Duvenaud, D.~Maclaurin, J.~Iparraguirre, R.~Bombarell, T.~Hirzel,
  A.~Aspuru-Guzik, and R.~P. Adams, ``Convolutional networks on graphs for
  learning molecular fingerprints,'' in \emph{Advances in Neural Information
  Processing Systems}, 2015, pp. 2224--2232.

\bibitem{defferrard2016convolutional}
M.~Defferrard, X.~Bresson, and P.~Vandergheynst, ``Convolutional neural
  networks on graphs with fast localized spectral filtering,'' in
  \emph{Advances in Neural Information Processing Systems}, 2016, pp.
  3844--3852.

\bibitem{hamilton2017inductive}
W.~Hamilton, Z.~Ying, and J.~Leskovec, ``Inductive representation learning on
  large graphs,'' in \emph{Advances in Neural Information Processing Systems},
  2017, pp. 1024--1034.

\bibitem{monti2017geometric}
F.~Monti, D.~Boscaini, J.~Masci, E.~Rodola, J.~Svoboda, and M.~M. Bronstein,
  ``Geometric deep learning on graphs and manifolds using mixture model
  {CNN}s,'' in \emph{IEEE Conference on Computer Vision and Pattern
  Recognition}, 2017, pp. 5115--5124.

\bibitem{kipf2018neural}
T.~Kipf, E.~Fetaya, K.-C. Wang, M.~Welling, and R.~Zemel, ``Neural relational
  inference for interacting systems,'' in \emph{International Conference on
  Machine Learning}, vol.~80, 2018, pp. 2693--2702.

\bibitem{velivckovic2017graph}
P.~Veli{\v{c}}kovi{\'c}, G.~Cucurull, A.~Casanova, A.~Romero, P.~Lio, and
  Y.~Bengio, ``Graph attention networks,'' \emph{arXiv preprint
  arXiv:1710.10903}, 2017.

\bibitem{li2018adaptive}
R.~Li, S.~Wang, F.~Zhu, and J.~Huang, ``Adaptive graph convolutional neural
  networks,'' in \emph{AAAI Conference on Artificial Intelligence}, 2018.

\bibitem{gao2018large}
H.~Gao, Z.~Wang, and S.~Ji, ``Large-scale learnable graph convolutional
  networks,'' in \emph{ACM SIGKDD International Conference on Knowledge
  Discovery \& Data Mining}, 2018, pp. 1416--1424.

\bibitem{liu2020refinedbox}
Y.~Liu, S.-J. Li, and M.-M. Cheng, ``Refinedbox: Refining for fewer and
  high-quality object proposals,'' \emph{Neurocomputing}, 2020.

\bibitem{li2020projected}
S.~Li, Y.~Liu, and J.~Gall, ``Projected-point-based segmentation: A new
  paradigm for lidar point cloud segmentation,'' \emph{arXiv preprint
  arXiv:2008.03928}, 2020.

\bibitem{cao2017realtime}
Z.~Cao, T.~Simon, S.-E. Wei, and Y.~Sheikh, ``Realtime multi-person 2d pose
  estimation using part affinity fields,'' in \emph{IEEE Conference on Computer
  Vision and Pattern Recognition}, 2017, pp. 7291--7299.

\bibitem{yeh2019chirality}
R.~Yeh, Y.-T. Hu, and A.~Schwing, ``Chirality nets for human pose regression,''
  in \emph{Advances in Neural Information Processing Systems}, 2019, pp.
  8161--8171.

\bibitem{du2015hierarchical}
Y.~Du, W.~Wang, and L.~Wang, ``Hierarchical recurrent neural network for
  skeleton based action recognition,'' in \emph{IEEE Conference on Computer
  Vision and Pattern Recognition}, 2015, pp. 1110--1118.

\bibitem{song2017end}
S.~Song, C.~Lan, J.~Xing, W.~Zeng, and J.~Liu, ``An end-to-end spatio-temporal
  attention model for human action recognition from skeleton data,'' in
  \emph{AAAI Conference on Artificial Intelligence}, 2017.

\bibitem{liu2017enhanced}
M.~Liu, H.~Liu, and C.~Chen, ``Enhanced skeleton visualization for view
  invariant human action recognition,'' \emph{Pattern Recognition}, vol.~68,
  pp. 346--362, 2017.

\bibitem{fan2020context}
Y.~Fan, S.~Weng, Y.~Zhang, B.~Shi, and Y.~Zhang, ``Context-aware
  cross-attention for skeleton-based human action recognition,'' \emph{IEEE
  Access}, vol.~8, pp. 15\,280--15\,290, 2020.

\bibitem{li2019learning}
Y.~Li, R.~Xia, X.~Liu, and Q.~Huang, ``Learning shape-motion representations
  from geometric algebra spatio-temporal model for skeleton-based action
  recognition,'' in \emph{IEEE International Conference on Multimedia and
  Expo}, 2019, pp. 1066--1071.

\bibitem{zhao2019bayesian}
R.~Zhao, K.~Wang, H.~Su, and Q.~Ji, ``Bayesian graph convolution lstm for
  skeleton based action recognition,'' in \emph{IEEE International Conference
  on Computer Vision}, 2019, pp. 6882--6892.

\end{thebibliography}

\end{document}